\documentclass[10pt,twocolumn,letterpaper]{article}

\usepackage{cvpr}              
\usepackage[accsupp]{axessibility}  
\usepackage{booktabs}

\usepackage{microtype}

\renewcommand{\paragraph}[1]{\vspace{.5em}\noindent\textbf{#1.}}

\usepackage[accsupp]{axessibility}
\usepackage{bbm}
\usepackage{amssymb,mathtools}
\usepackage{multirow}
\usepackage{colortbl}
\usepackage{soul}
\usepackage{comment}
\usepackage{makecell}
\usepackage{pifont}
\usepackage{wrapfig}
\usepackage{subcaption}
\usepackage{arydshln}

\usepackage[most]{tcolorbox}
\tcbuselibrary{listings, breakable, skins}

\definecolor{cvprblue}{rgb}{0.21,0.49,0.74}
\usepackage[pagebackref,breaklinks,colorlinks,allcolors=cvprblue]{hyperref}

\def\paperID{31232} 
\def\confName{CVPR}
\def\confYear{2026}

\title{TF-CADE: Foreground-Concentrated Text-Video Alignment for Zero-Shot Temporal Action Detection}

\newcommand{\authorskip}{\hspace{7.5mm}}
\newcommand{\customfootnote}[1]{%
  \begingroup
  \renewcommand{\thefootnote}{}
  \footnote{#1}%
  \endgroup
}

\author{
Yearang Lee \authorskip Ho-Joong Kim \authorskip
 Seong-Whan Lee$^*$ \\[2.5pt]
 \normalsize Dept. of Artificial Intelligence, Korea University, Seoul, Korea\\[-1pt]
 {\tt\small \{yr\_lee, hojoong\_kim, sw.lee\}@korea.ac.kr} \vspace{-6pt}\\
}

\begin{document}
\maketitle
\begin{abstract}
Zero-Shot Temporal Action Detection (ZSTAD) aims to localize and recognize action instances from unseen action categories in untrimmed videos. 
Although existing methods have shown effectiveness by advancing architectural text-video alignment, they still struggle with capturing semantic distinctions between action classes, resulting in text-irrelevant predictions.
To address this issue, we propose a Text-Foreground Concentrated Alignment for zero-shot temporal action DEtector (TF-CADE) that explicitly aligns textual information with action-relevant foreground regions.
Specifically, we introduce Action Concentrate Aggregation (ACA), which extracts action concentrate scores to aggregate temporally informative video segments into a foreground-weighted video embedding.
This foreground concentrated alignment enhances the semantic consistency between text and video features and improves inter-class discriminability.
In addition, a Certainty-based Confidence Re-weighting (CCR) strategy refines per-snippet confidence scores by leveraging foreground-aware similarity, effectively suppressing irrelevant action classes during inference.
Extensive evaluations show that our TF-CADE not only achieves state-of-the-art performance under in-distribution settings but also excels in cross-dataset generalization to unseen action classes.
\end{abstract}

\vspace{10pt}

\vspace{-22pt}
\customfootnote{*Corresponding author}
\section{Introduction}
\label{sec:intro}

    \begin{figure}[!t]
        \centering
        \begin{subfigure}[t]{0.87\linewidth}
            \centering
            \includegraphics[width=\linewidth]{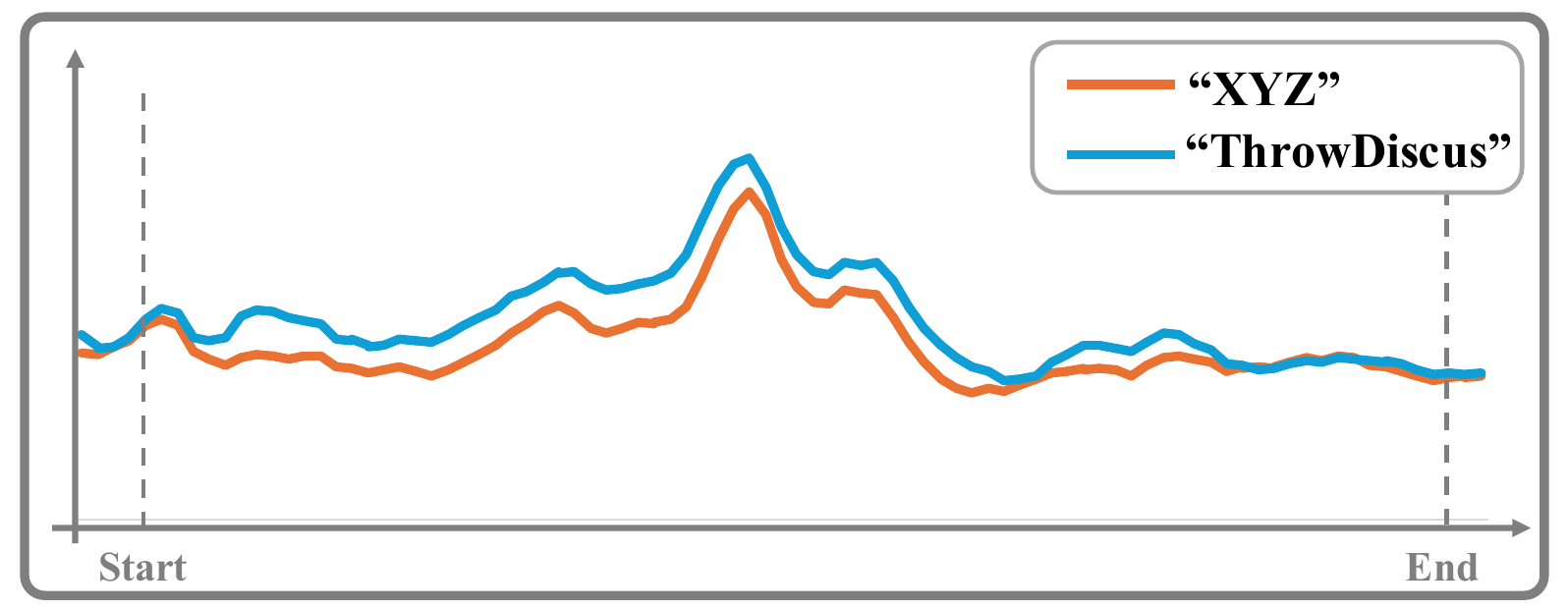}
            \caption{Ti-FAD~\cite{9TiFAD}}
        \end{subfigure}
        \\
        \begin{subfigure}[t]{0.87\linewidth}
            \centering
            \includegraphics[width=\linewidth]{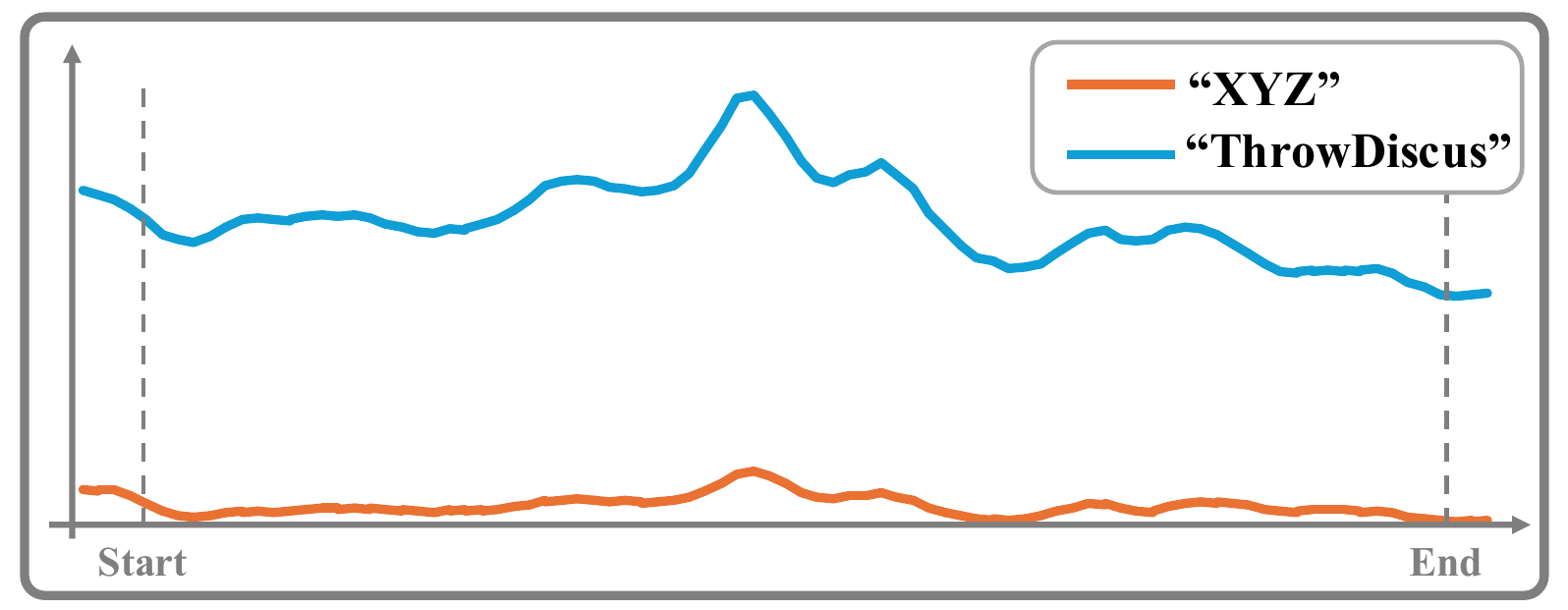} 
            \caption{TF-CADE (Ours)}
        \end{subfigure}
            
        \vspace{-3pt}
        \caption{
        \textbf{Class confidence scores under different textual inputs.}
        (a) Ti-FAD~\cite{9TiFAD} produces nearly identical confidence distributions whether the input text is the correct action label (``ThrowDiscus'') or an unrelated word (``XYZ''). In contrast, our (b) TF-CADE generates clearly differentiated confidence scores.
        }

        \label{fig:confidence_score}
        \vspace{-15pt}
    \end{figure}
    Zero-Shot Temporal Action Detection~(ZSTAD) aims to temporally localize and classify action instances from novel categories that are not seen during training time in untrimmed videos.
    With the advances of large-scale pre-trained visual-language models (e.g., CLIP \cite{CLIP} and ALIGN \cite{align}), existing ZSTAD methods \cite{3STALE,DeTAL,STOVTAL,6ZEETAD,mProTEA,OVFormer,9TiFAD,8T3AL} have leveraged their generalization capabilities by aligning the text features with the relevant temporal regions in the video. The existing ZSTAD methods are categorized into two types: foreground-based and foreground-free methods.
    
    The foreground-based methods~\cite{2EffPrompt,3STALE,DeTAL,STOVTAL,6ZEETAD} first extract foreground proposals to identify action candidate segments, and subsequently align these foreground proposals with text features.
    The pre-extracted foreground proposals are obtained independently without considering the text features, limiting the integration of textual representations.
    To overcome this limitation, existing foreground-free methods ~\cite{mProTEA,OVFormer,9TiFAD,8T3AL} incorporate text features directly into video features throughout the entire detection process. 
    Instead of relying on pre-extracted foreground proposals, the foreground-free methods leverage bi-directional cross-attention between text and video features, generating mutually enhanced representations.
    These mutually enhanced representations are subsequently utilized to produce class confidence scores during detection process.
    Among them, Ti-FAD~\cite{9TiFAD} introduces a cross-modal adaptation, in which text-infused bi-directional cross attention module aligns text with video features containing both foreground and background regions, producing text-aware video and video-aware text representations.
    However, despite such architectural improvements, the existing foreground-free methods such as Ti-FAD~\cite{9TiFAD} still struggle to incorporate text features into the detection process, producing text-irrelevant predictions.

    To demonstrate this issue, we analyze how the detector responds when conditioned on two distinct types of text inputs: a ground-truth action label and a meaningless word that has no semantic relation to any action category.
    Fig.~\ref{fig:confidence_score} compares the resulting class confidence scores computed via text-video similarity between the foreground-free method Ti-FAD~\cite{9TiFAD} and our method.
    As shown in Fig.~\ref{fig:confidence_score}~(a), Ti-FAD~\cite{9TiFAD} produces almost identical confidence distributions across all time steps, regardless of whether the input is the correct action name \emph{``ThrowDiscus''} or an irrelevant word such as \emph{``XYZ''}.
    This invariance of the confidence scores indicates that the existing foreground-free method fails to incorporate meaningful textual information, and struggles to distinguish between semantically valid and invalid text inputs, demonstrating that the prediction is predominantly driven by video features.
    This text-irrelevant prediction indicates that the foreground-free method does not incorporate meaningful guidance from textual inputs.

\begin{figure}[!t]
        \centering
        \begin{subfigure}[t]{0.49\linewidth}
            \centering
            \includegraphics[width=\linewidth]{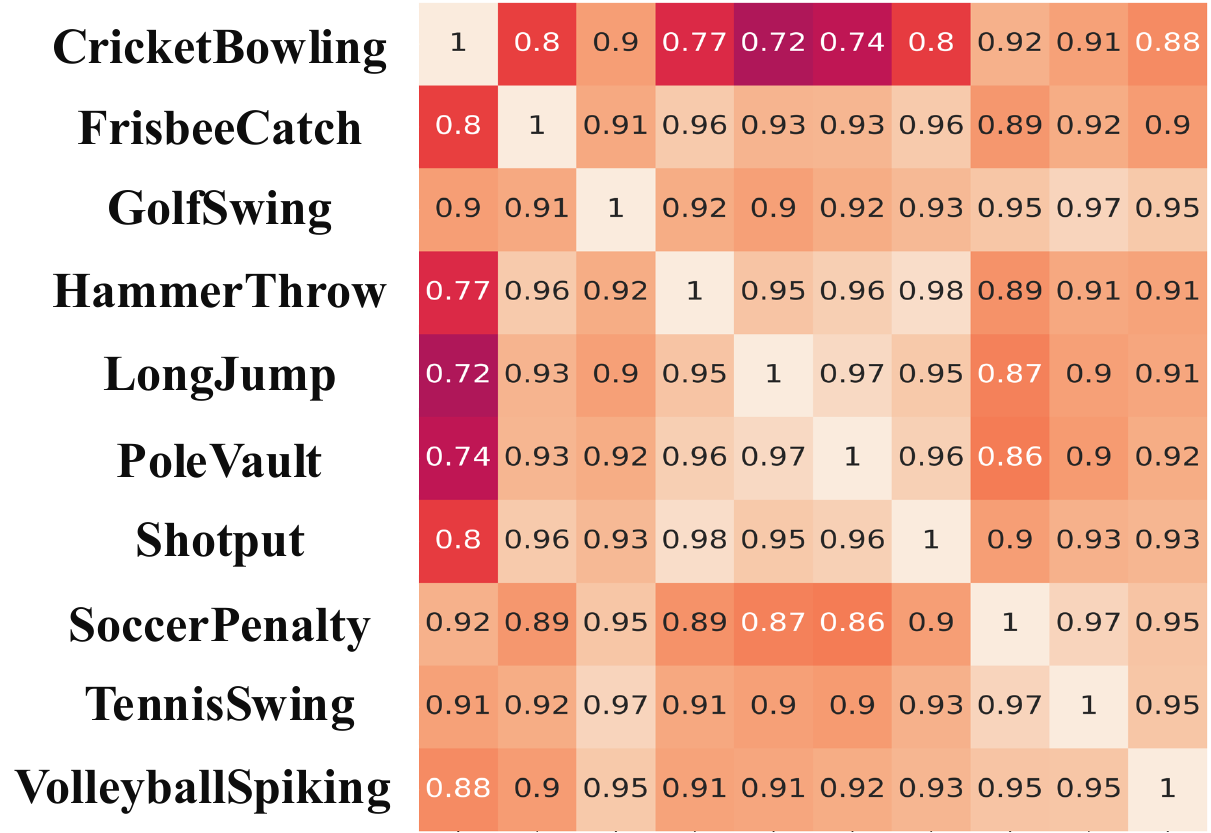}
            \caption{w/ foreground \& background}
        \end{subfigure}
        \hfill
        \begin{subfigure}[t]{0.49\linewidth}
            \centering
            \includegraphics[width=\linewidth]{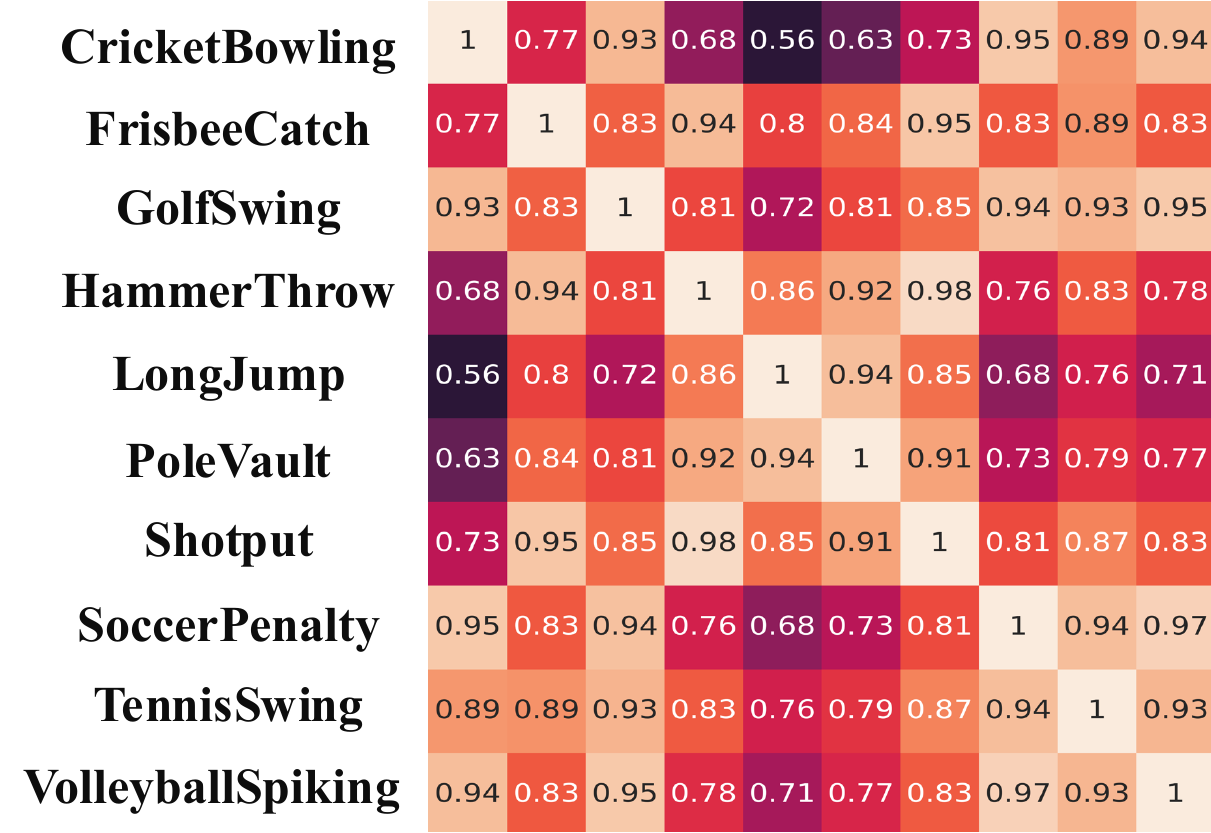}
            \caption{w/ foreground only}
        \end{subfigure}
        \hfill
        \vspace{-3pt}
        \caption{
        \textbf{Cosine similarity heatmap of the visual-aware text features produced by the cross-modal adaptation in Ti-FAD~\cite{9TiFAD} across various action classes on THUMOS14~\cite{THUMOS14}.}
        (a) With video features including both foreground and background regions, and 
        (b) with only foreground regions used for cross-modal adaptation.}
        
        \label{fig:fig2}
        \vspace{-15pt}
    \end{figure}

    To understand why textual cues fail to influence the prediction, we assume that the cross-modal adaptation in foreground-free method aligns text with video features that contain both action-relevant foreground and text-irrelevant background regions.
    Since background regions often dominate the visual representation in untrimmed videos, this joint alignment causes the updated text features to drift toward background-biased visual patterns.
    To examine this hypothesis, we analyze how the video-aware text features produced by the cross-modal adaptation in Ti-FAD~\cite{9TiFAD} represent when background influence is removed.
    Fig.~\ref{fig:fig2} visualizes the cosine similarity matrices of the resulting text features under two controlled conditions.
    As shown in Fig.~\ref{fig:fig2}~(a), the first condition reflects the standard foreground-free setting where all video regions are used for cross-modal adaptation.
    The resulting text features form uniformly high similarities across different action names, indicating that the model fails to encode class-specific textual cues.
    In contrast, Fig.~\ref{fig:fig2}~(b) shows the outcome when we construct a controlled variant that feeds only the ground-truth foreground regions, removing all background segments while keeping the overall detector architecture unchanged.
    This setup serves purely as a diagnostic experiment that removes the role of background during the alignment step.
    Under this foreground-only condition, the resulting text features become distinctly more separable across classes compared to Fig.~\ref{fig:fig2}(a).
    This comparison shows that background information interferes with the alignment between text features and action-related visual patterns.
    Consequently, these observations motivate the need for explicitly aligning text with action-relevant foreground regions, preventing background-dominated drift.
    
    
    


    In this paper, we propose a novel Text-Foreground Concentrated Alignment for zero-shot temporal action DEtector (TF-CADE), which explicitly aligns textual information with foreground regions that are semantically relevant to the actions. 
    To achieve this, we first introduce an Action Concentrate Aggregation (ACA) method that extracts  foreground-weighted video embedding by aggregating action concentrate scores over time, focusing on the text-relevant video regions.
    ACA employs a Gaussian smoothing mechanism to filter continuous action segments, thereby generating foreground-weighted video embedding that emphasizes temporally consistent action regions. 
    This foreground-weighted embedding is then used to produce a foreground-based video-level similarity score, which represents the overall action confidence of the video.
    During training, this foreground-weighted video embedding is aligned with its corresponding ground-truth text embedding.
    By selectively aggregating action-relevant regions, this alignment process enables the model to align text with action-relevant foreground regions.
    Additionally, we introduce a Certainty-based Confidence Re-weighting (CCR) strategy, which serves as a video-level prior during inference. 
    It applies the foreground-based video-level similarity scores to refine per-snippet classification confidence scores and suppress irrelevant action classes.
    Extensive experiments on standard in-distribution ZSTAD results show that TF-CADE outperforms state-of-the-art methods on THUMOS14 and ActivityNet~v1.3.
    Furthermore, our TF-CADE achieves superior performance in cross-dataset evaluation, demonstrating its strong generalization and zero-shot capability.

    \noindent In summary, our main contributions are as follows:
    \begin{itemize}[itemsep=1pt]
        \item We identify the key limitation of existing foreground-free ZSTAD models, which produce text-irrelevant predictions due to misalignment between text and video features.
        \item We propose TF-CADE with ACA to enhance video-text alignment between action-relevant segments and texts, and CCR to suppress confusing classes during inference.
        \item Extensive experiments across in-distribution and cross-dataset settings show TF-CADE achieves state-of-the-art performance on diverse datasets.
        
    \end{itemize}

\section{Related Work}
\label{sec:relatedwork}
    \noindent \textbf{Vision-language Models.}
    CoCa~\cite{coca}, CLIP~\cite{CLIP}, and ALIGN~\cite{align} are large-scale vision-language models trained with contrastive learning on massive image-text pairs, aiming to learn generalizable representations for open-set recognition tasks.
    These models align visual and textual modalities in a shared embedding space, enabling zero-shot transfer to a wide range of downstream tasks.
    Recent studies~\cite{relatedvlm1, relatedvlm2, relatedvlm3} have extended these models to video domains, demonstrating their strong potential for zero-shot temporal action understanding.
    In this work, we employ the visual and text encoders from CoCa, as well as the text encoder from CLIP, to extract the text features.
    
    \noindent \textbf{Temporal Action Detection.}
    Existing temporal action detection methods \cite{ActionFormer,Tridet,AFSD,TE-TAD} aim to localize and classify actions in untrimmed videos. 
    Temporal action detection is categorized into anchor-based and anchor-free approaches.
    Anchor-based methods rely on predefined temporal anchors or proposals to predict action boundaries and categories, which limits flexibility and increases computational cost due to dense anchor sampling.
    In contrast, anchor-free detectors directly predict action boundaries and confidence scores from time steps.
    Among them, ActionFormer~\cite{ActionFormer} introduce anchor-free detector, which directly predict action boundaries and classification scores from temporal feature sequences without the predefined temporal anchors.
    In this work, following the previous works~\cite{9TiFAD,DeTAL}, we adopt the anchor-free detector as our base architecture.
    
    \noindent \textbf{Zero-shot Temporal Action Detection.}
    ZSTAD aims to detect and classify previously unseen action classes in untrimmed videos.
    Existing ZSTAD approaches can be categorized into two types: foreground-based~\cite{2EffPrompt,3STALE,DeTAL,STOVTAL,6ZEETAD} and foreground-free methods~\cite{mProTEA,OVFormer,9TiFAD,8T3AL,ZEAL}.
    The foreground-based methods first identify foreground proposals without incorporating text features, and subsequently align these pre-extracted proposals with text embeddings.
    STALE~\cite{3STALE} adopts a masking strategy to extract foreground proposals but does not incorporate text features in the extraction process.
    Ti-FAD~\cite{9TiFAD} introduces a foreground-free framework that jointly aligns video and text features via text-infused attention.
    Although Ti-FAD~\cite{9TiFAD} achieves strong performance through architectural improvements, they still struggle to integrate text features into the detection process.
    Recently, T3AL~\cite{8T3AL} proposes a training-free method that uses a text-guided region suppression mechanism to align text inputs with meaningful video features.
    However, T3AL relies on an additional captioning model (e.g., CoCa~\cite{coca}) to generate sentence-level descriptions, making the pipeline dependent on external language models.
    In contrast, our approach focuses on explicitly aligning meaningful visual features with text embeddings without relying on any captioning model or auxiliary classification scores (e.g., from a video-level classifier such as UntrimmedNets~\cite{untrimmednets}).

\section{Preliminary}

    \subsection{Problem Setting}
        Let \( X = \{x_t\}_{t=1}^{T_0} \) denote the sequence of snippet-level features from an untrimmed video, where \( T_0 \) is the number of snippets that corresponds to a few sequence of consecutive frames. 
        These snippets are processed using a pre-trained visual backbone (e.g., I3D~\cite{I3D}, InternVideo2~\cite{internvideo2}). 
        Each video is annotated with \( N_a \) action instances \( \mathcal{A} = \{ (s_a, e_a, y_a) \}_{a=1}^{N_a} \),  
        where \( s_a \) and \( e_a \) denote the start and end times of the \( a \)-th action instance, and \( y_a \in \mathcal{C} \) denotes its class name from the set of all action names \( \mathcal{C} \), with \( |\mathcal{C}| = N_c \).
        In the zero-shot setting, the training and test action categories, denoted as \( S_{\text{train}} \) and \( S_{\text{test}} \), are disjoint, i.e., \( S_{\text{train}} \cap S_{\text{test}} = \emptyset \). ZSTAD addresses the problem of detecting actions from unseen categories that do not appear in the training data.

    
    \subsection{Cross-modal Adaptation Baseline}
        \label{sec:baseline}
        Ti-FAD~\cite{9TiFAD} introduces a cross-modal adaptation baseline designed to enhance the alignment between text and video features. 
        The snippet-level features $X$ are projected into initial video embeddings  \( v^{(0)} = \{ v_t^{(0)} \}_{t=1}^{T_0} \in \mathbb{R}^{T_0 \times D} \) via 1D convolution layers, where $D$ is the embedding dimension of the detector.
        The initial text embeddings \( c^{(0)} = \{ c_n^{(0)} \}_{n=1}^{N_c} \in \mathbb{R}^{N_c \times D} \) are obtained using a pre-trained text encoder such as CLIP~\cite{CLIP}.
        These initial video \( v^{(0)} \) and text features \( c^{(0)} \) are updated through an encoder module $\text{Encoder}^{(l)}(\cdot,\cdot)$, which consists of self-attention layers for video and text, cross-attention between the two modalities, and feed-forward networks for each modality, applied at each layer $l$.
        After the self-attention, the video features are temporally downsampled to produce multi-scale representations \( v^{(l)} = \{ v_t^{(l)} \}_{t=1}^{T_l} \), where \( T_l = T_{l-1} / 2 \) is the temporal length at $l$.
        Meanwhile, the text features are progressively updated at each level as \( c^{(l)} = \{ c_n^{(l)} \}_{n=1}^{N_c} \).
        At each level, cross-attention is performed with video features as queries to attend to the text features, and the outputs are refined via feed-forward layers.
        The $l$-th encoded features are represented as follows:
        \begin{equation}
            v^{(l)}, c^{(l)} = \text{Encoder}^{(l)}(v^{(l-1)}, c^{(l-1)}) \quad l = 1, \dots, L\text{,}
        \end{equation}
        where $L$ denotes the total number of pyramid levels.


\begin{figure*}[!t]
        \centering
        \begin{subfigure}[t]{0.49\linewidth}
            \centering
            \includegraphics[width=\linewidth]{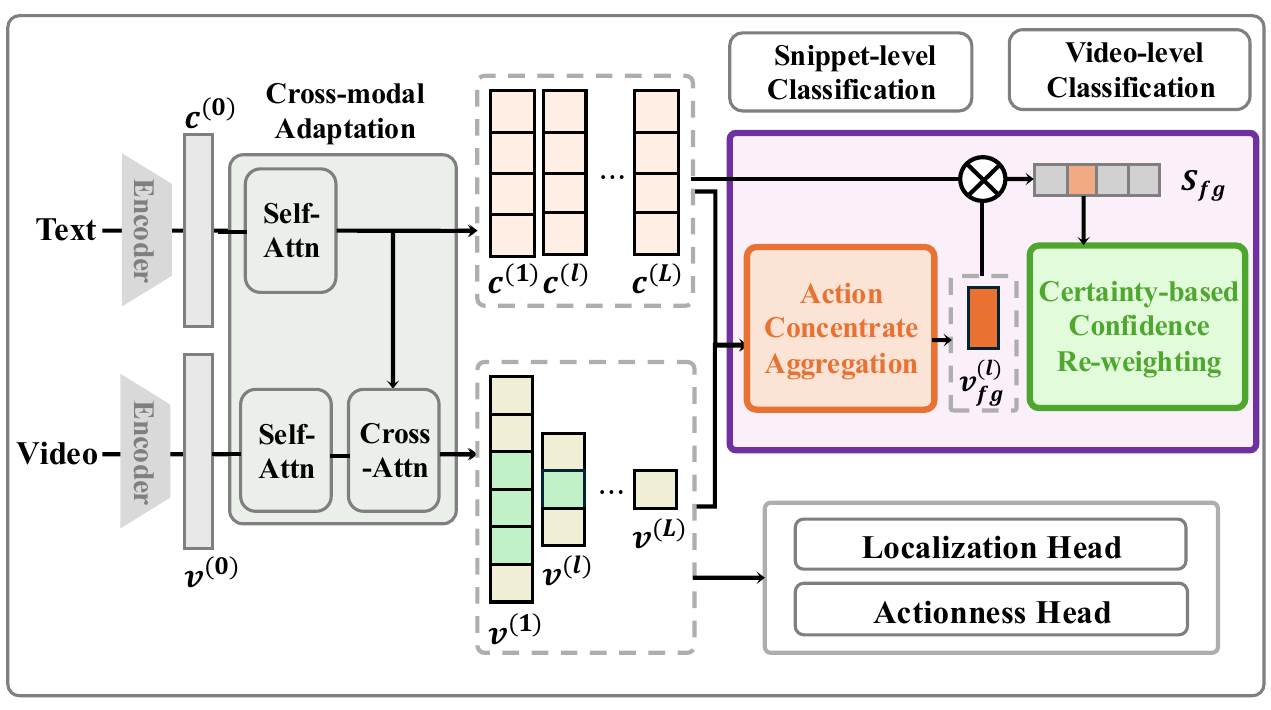}
        \end{subfigure}
        \begin{subfigure}[t]{0.5\linewidth}
            \centering
            \includegraphics[width=\linewidth]{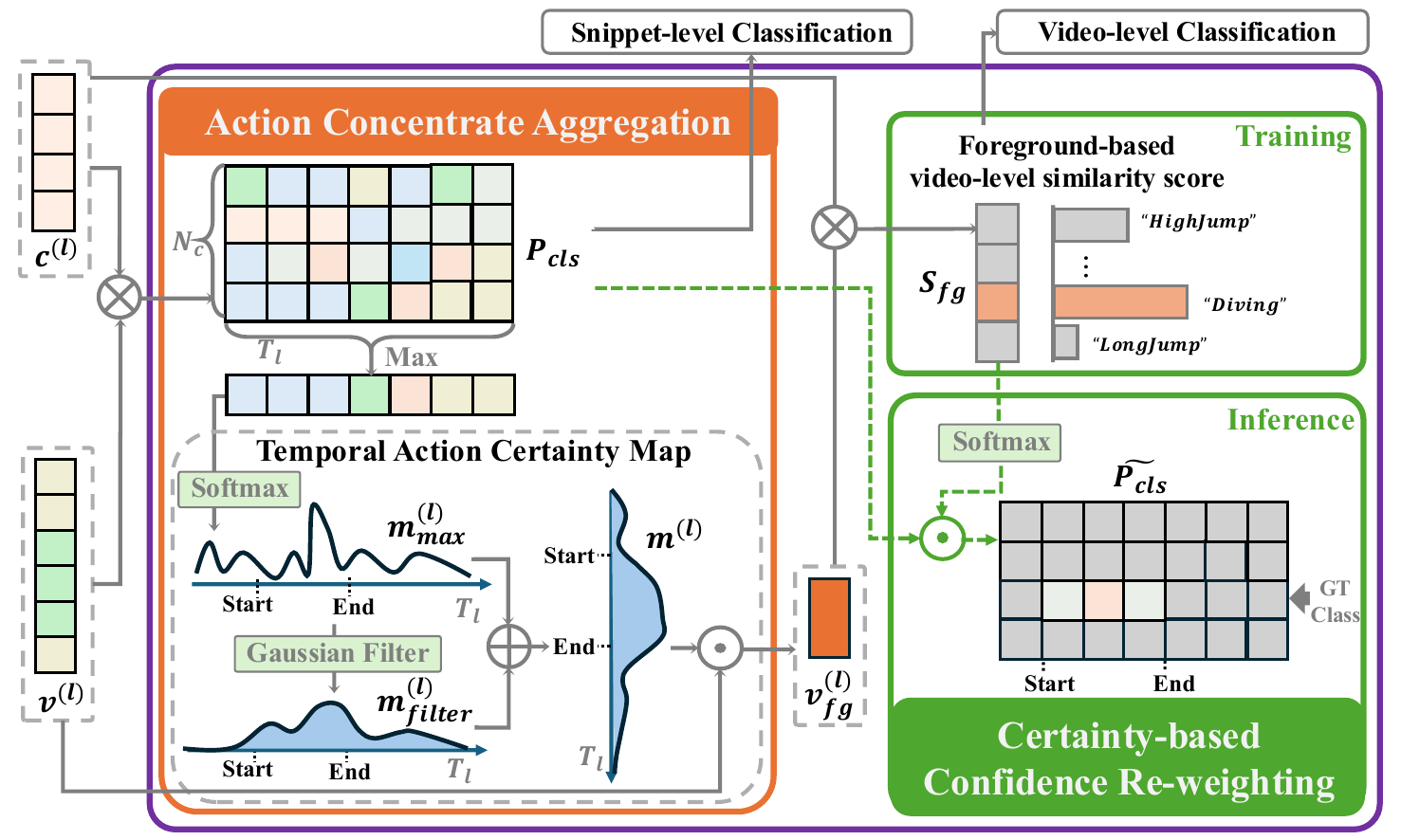}
        \end{subfigure}
        \caption{
        \textbf{Overview of TF-CADE.} 
        (a) Overall framework of the proposed TF-CADE, which builds upon the cross-modal adaptation baseline and introduces text-foreground concentrated alignment to explicitly associate textual information with action-relevant visual regions.
        (b) Illustration of the text-foreground concentrated alignment method at the \( l \)-th layer, which consists of ACA and CCR.
        ACA computes temporal action certainty scores and generates a foreground-weighted video embedding for foreground-text alignment during training, while CCR refines per-snippet classification scores using the foreground-based video-level similarity score during inference.}
        
        \label{fig:overview}
        \vspace{-11pt}
    \end{figure*}

\section{TF-CADE}

\subsection{Overview}
    In this section, we describe our TF-CADE, which aligns text features with foreground-concentrated visual features to mitigate the influence of background regions containing action-irrelevant visual patterns.
    Built upon a cross-modal adaptation baseline following Ti-FAD~\cite{9TiFAD}, our TF-CADE is composed of two main components:
    (1) Action Concentrate Aggregation (ACA), which computes temporal action certainty scores and aggregates video features into a foreground-weighted video embedding for text-foreground alignment; and
    (2) Certainty-based Confidence Re-weighting (CCR), which refines snippet-level classification logits during inference by incorporating the foreground-based video-level similarity scores, thereby suppressing irrelevant action classes.
    An overview of TF-CADE is illustrated in Fig.~\ref{fig:overview}.
    
    \subsection{Action Concentrate Aggregation}
        To align text with foreground visual features, we introduce ACA: (1) obtaining a temporal action certainty map applying smoothing strategy with Gaussian filter; (2) producing a video-level similarity score aggregating video features weighted by foreground information and comparing them with candidate class embeddings.
        
        \noindent \textbf{Temporal Action Certainty Map.}
        To obtain action-relevant foreground visual features, we construct temporal action certainty map. 
        We utilize the multi-level video features \( v^{(l)}\) and text features \( c^{(l)} \) at each level \( l = 1, \dots, L, \) of the $\text{Encoder}(\cdot)$ output.
        We obtain a video-text confidence score map $P_{\text{cls}}$ via similarity between video and text features:
        \begin{equation}
            \label{equ:p_cls}
            P_{\text{cls}} = v^{(l)} \cdot {c^{(l)}}^\top \in \mathbb{R}^{T_l \times N_c}.
        \end{equation}
        To obtain an action certainty score, we first take the maximum similarity across the class dimension and apply a softmax function as follows:
        \begin{equation}
            m_{\text{max}}^{(l)} = \text{softmax}\left( \max_{N_{c}} \left( P_{\text{cls}} \right) \right) \in \mathbb{R}^{T_l}\text{,}
        \end{equation}
        where $m_{\text{max}}^{(l)}$ represents the initial action certainty at level $l$.
        $m_{\text{max}}^{(l)}$ is sharply concentrated on the most salient action frames via the $\text{softmax}$ function, representing the peak occurrence of the action.

        \noindent To aggregate the full temporal context over the entire foreground action segment, we apply a smoothing method to $m_{\text{max}}^{(l)}$. 
        Specifically, we generate a smoothed action certainty map $m_{\text{filter}}^{(l)}$ by applying Gaussian kernel $G(\sigma)$ along the temporal dimension $T_{l}$ as follows:
        \begin{equation}
            m_{\text{filter}}^{(l)} = m_{\text{max}}^{(l)} \circledast G(\sigma) \in \mathbb{R}^{T_l}\text{,}
        \end{equation}
        where $\circledast$ represents the 1D convolution operation, and $\sigma$ denotes the standard deviation of the Gaussian kernel that controls the degree of temporal smoothing.
        This smoothing process reduces impact of noise and overly sharp peaks, promoting more stable temporal pooling.
        The final temporal action certainty map $m^{(l)}$ is obtained by combining the initial sharp saliency and the smoothed contextual saliency:
        \begin{equation}
            m^{(l)} = m_{\text{max}}^{(l)} + m_{\text{filter}}^{(l)}.
        \end{equation}
        We then normalize $m^{(l)}$ along the temporal dimension by dividing it by the sum of all time-step values, ensuring a consistent scale across the sequence.
        

        \noindent \textbf{Foreground-weighted Video Embedding.}
        \label{sec:videoscore}
        Utilizing the action certainty \( m^{(l)} \), we compute the foreground-weighted video embedding \( v_{\text{fg}}^{(l)} \) from the 
        video features via soft aggregation as follows:
        \begin{equation}
            v_{\text{fg}}^{(l)} = \sum_{t=1}^{T_l} m_t^{(l)} \odot v_t^{(l)} \in \mathbb{R}^{D}\text{,}
        \end{equation}
        where $\odot$ represents the element-wise product and $v^{(l)}$ denotes $l$-th video features produced by cross-modal adaptation module.
        We then calculate the foreground-based video-level score using the foreground-weighted video embedding $v_{\text{fg}}^{(l)}$ and text embedding $ c_n^{(l)}$.
        The temporal foreground-based video-level score ${S^{(n)}_{\text{fg}}}$ for the \( n \)-th class is calculated by averaging the similarities across all \( L \) levels:
        \begin{equation}
        \label{eq:video_level_score}
        {S^{(n)}_{\text{fg}}} = \frac{1}{L} \sum_{l=1}^{L} \text{sim}(v_{\text{fg}}^{(l)},  c_n^{(l)}) \quad n = 1, \dots, N_c\text{,}
        \end{equation}
        where $\text{sim}(\cdot,\cdot)$ denotes the cosine similarity.
        We denote the final similarity vector as ${S_{\text{fg}}} = \{{S^{(n)}_{\text{fg}}}\}_{n=1}^{N_c}$, which represents the foreground-based video-level similarity scores used for text-foreground concentrated alignment.

    \subsection{Certainty-based Confidence Re-weighting}
        \label{sec:filtering}
        In the standard ZSTAD inference pipeline, the video-text confidence score map defined in Eq.~\eqref{equ:p_cls} is passed through a sigmoid function to produce per-snippet classification scores.
        However, this snippet-level prediction often leads to over-activation of visually similar but semantically irrelevant classes. 
        To suppress confusing classes that are semantically similar to the target class, we introduce CCR that refines the confidence distribution using a video-level prior.
        CCR leverages the foreground-based video-level similarity score $S_{\text{fg}}$, obtained via action concentrate aggregation in Eq. \eqref{eq:video_level_score}.
        We apply a softmax function to $S_{\text{fg}}$ to estimate the likelihood of each class being present in the video.
        We then re-weight the snippet-wise class confidence map by performing element-wise multiplication with $S_{\text{fg}}$ as prior:
        \begin{equation}
            \label{eq:scr}
            \tilde{P}_{\text{cls}} = \sqrt{\text{sigmoid}(P_{\text{cls}}) \odot \text{softmax}({S_{\text{fg}}})} \in \mathbb{R}^{T_l \times N_c}\text{,}
        \end{equation}
        where ${\tilde{P}_{\text{cls}}}$ represents final re-weighted class confidence score map and $\odot$ represents the element-wise product.
        This re-weighting emphasizes action-relevant classes while suppressing irrelevant ones.

    \subsection{Training and Inference}
        \noindent \textbf{Training.}
        The classification loss is defined as:
        \begin{equation}
            \label{eq:cls_loss}
            \mathcal{L}_{cls} = \mathcal{L}_{snippet} + \mathcal{L}_{video},
        \end{equation}
        where $\mathcal{L}_{snippet}$ supervises snippet-level classification based on the class confidence map in Eq. \eqref{equ:p_cls}, 
        and $\mathcal{L}_{video}$ aligns the foreground-based video-level similarity score $S_{\text{fg}}$ with the corresponding action class.
        Both losses are computed using the focal loss~\cite{focalloss}.
        Following the previous methods~\cite{9TiFAD, ActionFormer}, we define the total objective function described as:
        $ \mathcal{L} = \mathcal{L}_{cls} + \mathcal{L}_{loc}+\mathcal{L}_{an}$, where localization loss $\mathcal{L}_{loc}$ regresses action boundaries using the DIoU loss~\cite{DIoUloss}, and $\mathcal{L}_{an}$ is actioness loss~\cite{ActionFormer} computed using the focal loss~\cite{focalloss}.
    
        \noindent \textbf{Inference.}
        For each time step \( t \), the model predicts $(\hat{s}, \hat{e}, \hat{p}_{cls})$, where $\hat{s}$ and $ \hat{e}$ represent the estimated start and end points of action instance, and $\hat{p}_{cls}$ denotes the corresponding action confidence score by applying the argmax over the final filtered class confidence score map $\tilde{P}_{\text{cls}}$ in Eq. \eqref{eq:scr}.
        Subsequently, we apply Soft-NMS~\cite{softnms} to suppress redundant proposals and retain the most confident detections.

\section{Experiments}
    \subsection{Experiment Settings}
    \label{sec:experiment_settings}
        \noindent \textbf{Datasets.}
        We use three datasets: THUMOS14~\cite{THUMOS14}, ActivityNet~v1.3~\cite{ActivityNet}, and HACS-Segment~\cite{hacs}.
        THUMOS14 contains 20 action classes related to sports across 200 training and 213 test videos.
        ActivityNet~v1.3 includes 200 action classes covering daily actions and contains 19,994 videos.
        HACS-Segment is a large-scale human activities dataset consisting of 50,000 videos across 200 action classes. 
        We consider split settings for zero-shot scenario: 50\%-50\% setting (training on 50\% of the classes and testing on the remaining 50\%) and 75\%-25\% setting (training on 75\% of the classes and testing on the remaining 25\%) following EffPrompt~\cite{2EffPrompt}.
        
        \noindent \textbf{Evaluation Protocol and Comparison Methods.}
        To ensure a strictly zero-shot evaluation for ZSTAD, we focus our comparison on models that do not rely on external classification scores.
        The external classification scores are commonly used as a post-processing step, obtained from a supervised classifier such as UntrimmedNet~\cite{untrimmednets}, but cannot address novel classes because they are trained in closed-set supervised setting. 
        Although most existing ZSTAD methods report impressive results, they rely on such external classifiers, introducing additional supervision that violates zero-shot assumption.
        Therefore, we select following models that have publicly available implementations, removing only the external classification post-processing without modifying any other components: a foreground-based method STALE~\cite{3STALE}; training-free method T3AL~\cite{8T3AL}; and a state-of-the-art foreground-free method Ti-FAD~\cite{9TiFAD}.

        \begin{table*}[!ht]
\setlength{\tabcolsep}{5pt}
\caption{\textbf{Performance comparison with the state-of-the-art methods on THUMOS14 and ActivityNet v1.3 under in-distribution setting.} The average mAP is calculated to evaluate performance at different tIoU thresholds: [0.3:0.1:0.7] for THUMOS14 and [0.5:0.05:0.95] for ActivityNet~v1.3. $^\dagger$ indicates that the results are produced utilizing additional text descriptions (e.g., sentence-level captions by CoCa~\cite{coca}).}
\vspace{-3pt}
\centering
\renewcommand{\arraystretch}{1.1}
\resizebox{0.79\linewidth}{!}{
\begin{tabular}{clcccccccccccccc}
\toprule[1.5pt]
\multirow{2.5}{*}{Train Split} 
& \multicolumn{1}{l}{\multirow{2.5}{*}{Model}} 
& \multicolumn{2}{c}{Feature} & 
& \multicolumn{6}{c}{THUMOS14} 
& & \multicolumn{4}{c}{ActivityNet~v1.3} \\
\noalign{\smallskip}
\cline{3-4}
\cline{6-11}
\cline{13-16}
\noalign{\smallskip}
 & & Video & Text & & 0.3 & 0.4 & 0.5 & 0.6 & 0.7 & Avg. & & 0.5 & 0.75 & 0.95 & Avg.\\
\noalign{\smallskip}
\hline
\hline
\noalign{\smallskip}
\multirow{17}{*}{\shortstack{50\% Seen\\ 50\% Unseen}} 
& EffPrompt \cite{2EffPrompt} & I3D \cite{I3D} & CLIP-B \cite{CLIP} & & 37.2 & 29.6 & 21.6 & 14.0 & 7.2 & 21.9 & & 32.0 & 19.3 & 2.9 & 19.6 \\
& STALE \cite{3STALE}  & I3D \cite{I3D} & CLIP-B \cite{CLIP} & & 38.3 & 30.7 & 21.2 & 13.8 & 7.0 & 22.2 & & 32.1 & 20.7 & \underline{5.9} & 20.5 \\
& DeTAL \cite{DeTAL} & I3D~\cite{I3D} / R(2+1)D~\cite{R21D} & CLIP-B \cite{CLIP} & &  38.3 & 32.3 & 24.4 & 16.3 & 9.0 & 24.1 & & 34.4 & 23.0 & 4.0 & 22.4 \\
& STOV-TAL \cite{STOVTAL} & CLIP-B \cite{CLIP} & CLIP-B \cite{CLIP} & & 44.2 & 35.7 & 25.7 & 16.5 & 8.0 & 26.0 & & 42.1 & 25.0 & 1.3 & 24.8 \\
& ZEETAD \cite{6ZEETAD} & I3D \cite{I3D} & CLIP-B \cite{CLIP} & & 45.2 & 38.8 & 30.8 & 22.5 & 13.7 & 30.2 & & 39.2 & 25.7 & 3.1 & 24.9 \\
\noalign{\smallskip}
\cline{2-16}
\noalign{\smallskip}
& mProTEA \cite{mProTEA} & CLIP-B \cite{CLIP} & CLIP-B \cite{CLIP} & & 41.2 & 36.3 & 26.3 & 16.8 & 8.4 & 26.1 & & 41.8 & 24.6 & 6.1 & 25.6 \\
& OVFormer \cite{OVFormer} & I3D \cite{I3D} & CLIP-B \cite{CLIP} & & 42.8$^\dagger$ & 37.3$^\dagger$ & 30.6$^\dagger$ & 23.5$^\dagger$ & 15.9$^\dagger$ & 30.5$^\dagger$ & & 42.8$^\dagger$ & 27.3$^\dagger$ & 6.0$^\dagger$ & 27.2$^\dagger$ \\
& Ti-FAD \cite{9TiFAD} & I3D \cite{I3D} & CLIP-B \cite{CLIP} & & \underline{57.0} & \underline{51.4} & \underline{43.3} & \underline{33.0} & \underline{21.2} & \underline{41.2} & & \underline{50.6} & \underline{32.2} & 5.2& \underline{32.0} \\
\noalign{\smallskip}
\cline{2-16}
\noalign{\smallskip}
& \multicolumn{11}{l}{\textit{No external information}} && \\
& STALE \cite{3STALE} & I3D \cite{I3D} & CLIP-B \cite{CLIP} & & - & - & - & - & - & -  & & 6.9 & 4.7 & 0.9 & 4.4 \\
& T3AL \cite{8T3AL} & CoCA~\cite{coca} & CoCA~\cite{coca} &  & 20.7$^\dagger$ & 14.3$^\dagger$ & 8.9$^\dagger$ & 5.3$^\dagger$ & 2.7$^\dagger$ & 10.4$^\dagger$ & & 25.8$^\dagger$ & 13.9$^\dagger$ & 3.1$^\dagger$ & 14.3$^\dagger$ \\
& ZEAL \cite{ZEAL} & LLaVA-OneVision \cite{llava} & CLIP-L \cite{CLIP} & & 21.1 &15.0 &9.9& 5.0 &2.6 &10.7 & & - & - & - & - \\ 
& Ti-FAD \cite{9TiFAD} & CoCA~\cite{coca} & CoCA~\cite{coca} & & 15.4 & 13.6 & 11.1 & 8.3 & 5.3 & 10.7 & & 12.3 & 9.0 & \textbf{2.3} & 8.6 \\
& Ti-FAD \cite{9TiFAD} & I3D \cite{I3D} & CLIP-B \cite{CLIP} & & 20.4 & 19.1 & 17.0 & 13.8 & 9.6 & 16.0 & & 10.6 & 7.8 & 1.9 & 7.4 \\
& Ti-FAD \cite{9TiFAD} + Ours   & I3D \cite{I3D} & CLIP-B \cite{CLIP} & & 24.9 & 23.0 & 20.1 & 15.9 & 10.8 & 18.9 & & 14.7 & 9.2 & 1.4 & 9.2 \\
& \cellcolor[gray]{0.92}\textbf{TF-CADE (Ours)} & \cellcolor[gray]{0.92}CoCA~\cite{coca} & \cellcolor[gray]{0.92}CoCA~\cite{coca} & \cellcolor[gray]{0.92} & \cellcolor[gray]{0.92}19.7 & \cellcolor[gray]{0.92}17.1 & \cellcolor[gray]{0.92}13.1 & \cellcolor[gray]{0.92}9.3 & \cellcolor[gray]{0.92}5.5 & \cellcolor[gray]{0.92}12.9 & \cellcolor[gray]{0.92} & \cellcolor[gray]{0.92}\textbf{20.8} & \cellcolor[gray]{0.92}\textbf{13.1} & \cellcolor[gray]{0.92}2.1 & \cellcolor[gray]{0.92}\textbf{13.0} \\
& \cellcolor[gray]{0.92}\textbf{TF-CADE (Ours)} & \cellcolor[gray]{0.92}I3D \cite{I3D} & \cellcolor[gray]{0.92}CLIP-B \cite{CLIP} & \cellcolor[gray]{0.92} & \cellcolor[gray]{0.92}\textbf{28.6} & \cellcolor[gray]{0.92}\textbf{26.2} & \cellcolor[gray]{0.92}\textbf{22.6} & \cellcolor[gray]{0.92}\textbf{17.1} & \cellcolor[gray]{0.92}\textbf{10.7} & \cellcolor[gray]{0.92}\textbf{21.1} & \cellcolor[gray]{0.92} & \cellcolor[gray]{0.92}16.7 & \cellcolor[gray]{0.92}10.6 & \cellcolor[gray]{0.92}1.9 & \cellcolor[gray]{0.92}10.5\\
\noalign{\smallskip}
\hline
\noalign{\smallskip}
\multirow{17}{*}{\shortstack{75\% Seen\\ 25\% Unseen}}
& EffPrompt \cite{2EffPrompt} & I3D \cite{I3D} & CLIP-B \cite{CLIP} & & 39.7 & 31.6 & 23.0 & 14.9 & 7.5 & 23.3 & & 37.6 & 22.9 & 3.8 & 23.1 \\
& STALE \cite{3STALE} & I3D \cite{I3D} & CLIP-B \cite{CLIP} & & 40.5 & 32.3 & 23.5 & 15.3 & 7.6 & 23.8 & & 38.2 & 25.2 & 6.0 & 24.9 \\
& DeTAL \cite{DeTAL} & I3D~\cite{I3D} / R(2+1)D~\cite{R21D} & CLIP-B \cite{CLIP} & & 39.8 & 33.6 & 25.9 & 17.4 & 9.9 & 25.3 & & 39.3 & 26.4 & 5.0 & 25.8 \\
& STOV-TAL \cite{STOVTAL} & CLIP-B \cite{CLIP} & CLIP-B \cite{CLIP} & & 47.8 & 39.1 & 28.4 & 17.6 & 9.1 & 28.4 & & 47.0 & 28.1 & 1.6 & 27.9 \\
& ZEETAD \cite{6ZEETAD} & I3D \cite{I3D} & CLIP-B \cite{CLIP} & & 61.4 & 53.9 & 44.7 & 34.5 & 20.5 & 43.2 & & 51.0 & 33.4 & 5.9 & 32.5 \\
\noalign{\smallskip}
\cline{2-16}
\noalign{\smallskip}
& mProTEA \cite{mProTEA} & CLIP-B \cite{CLIP} & CLIP-B \cite{CLIP} & & 43.1 & 38.2 & 28.2 & 18.1 & 8.7 & 27.9 & & 44.5 & 27.4 & \underline{7.9} & 27.6 \\
& OVFormer \cite{OVFormer} & I3D \cite{I3D} & CLIP-B \cite{CLIP} & & 49.8$^\dagger$ & 43.8$^\dagger$ & 35.8$^\dagger$ & 27.8$^\dagger$ & 19.2$^\dagger$ & 35.3$^\dagger$ & & 46.7$^\dagger$ & 29.4$^\dagger$ & 6.1$^\dagger$ & 29.5$^\dagger$ \\
& Ti-FAD \cite{9TiFAD} & I3D \cite{I3D} & CLIP-B \cite{CLIP} & & \underline{64.0} & \underline{58.5} & \underline{49.7} & \underline{37.7} & \underline{24.1} & \underline{46.8} & & \underline{53.8} & \underline{34.8} & 7.0& \underline{34.7}\\
\noalign{\smallskip}
\cline{2-16}
\noalign{\smallskip}
& \multicolumn{11}{l}{\textit{No external information}} && \\
& STALE \cite{3STALE} & I3D \cite{I3D} & CLIP-B \cite{CLIP} & & - & - & - & - & - & - & & 14.0 & 9.8 & 2.0 & 9.5  \\
& T3AL \cite{8T3AL} & CoCA~\cite{coca} & CoCA~\cite{coca} & & 19.2$^\dagger$ & 12.7$^\dagger$ & 7.4$^\dagger$ & 4.4$^\dagger$ & 2.2$^\dagger$ & 9.2$^\dagger$ & & 28.1$^\dagger$ & 14.9$^\dagger$ & 3.3$^\dagger$ & 15.4$^\dagger$ \\
& ZEAL \cite{ZEAL} & LLaVA-OneVision \cite{llava} & CLIP-L \cite{CLIP} & & 22.1 &16.1& 11.0 &5.7 &3.0& 11.6 & & - & - & - & - \\ 
& Ti-FAD \cite{9TiFAD} & CoCA~\cite{coca} & CoCA~\cite{coca} & & 17.1 & 15.2 & 12.4 & 9.2 & 5.9 & 12.0 & & 22.6 & 15.7 & \textbf{3.8} & 15.3 \\
& Ti-FAD \cite{9TiFAD} & I3D \cite{I3D} & CLIP-B \cite{CLIP} & & 34.9 & 32.3 & 28.3 & 23.0 & 16.0 & 26.9 & & 19.6 & 14.2 & 3.5 & 13.7 \\
& Ti-FAD \cite{9TiFAD} + Ours & I3D \cite{I3D} & CLIP-B \cite{CLIP} & & 42.4 & 38.7 & 33.1 & 25.7 & 16.4 & 31.2 & & 25.6 & 16.5 & 3.2 & 16.4\\
& \cellcolor[gray]{0.92}\textbf{TF-CADE (Ours)} & \cellcolor[gray]{0.92}CoCA~\cite{coca} & \cellcolor[gray]{0.92}CoCA~\cite{coca} & \cellcolor[gray]{0.92} & \cellcolor[gray]{0.92}33.3 & \cellcolor[gray]{0.92}28.4 & \cellcolor[gray]{0.92}21.6 & \cellcolor[gray]{0.92}14.4 & \cellcolor[gray]{0.92}7.9 & \cellcolor[gray]{0.92}21.1 & \cellcolor[gray]{0.92} & \cellcolor[gray]{0.92}\textbf{30.3} & \cellcolor[gray]{0.92}\textbf{17.8} & \cellcolor[gray]{0.92}2.7 & \cellcolor[gray]{0.92}\textbf{18.0} \\
& \cellcolor[gray]{0.92}\textbf{TF-CADE (Ours)} & \cellcolor[gray]{0.92}I3D \cite{I3D} & \cellcolor[gray]{0.92}CLIP-B \cite{CLIP} & \cellcolor[gray]{0.92} & \cellcolor[gray]{0.92}\textbf{47.7} & \cellcolor[gray]{0.92}\textbf{43.2} & \cellcolor[gray]{0.92}\textbf{36.7} & \cellcolor[gray]{0.92}\textbf{27.9} & \cellcolor[gray]{0.92}\textbf{16.9} & \cellcolor[gray]{0.92}\textbf{34.5} & \cellcolor[gray]{0.92} & \cellcolor[gray]{0.92}27.4 & \cellcolor[gray]{0.92}17.1 & \cellcolor[gray]{0.92}3.2 & \cellcolor[gray]{0.92}17.2\\
\bottomrule[1.5pt]
\end{tabular}}
\vspace{-3pt}
\label{tab:thumos}
\vspace{-11pt}
\end{table*}

        \noindent \textbf{Evaluation Metric.}
            We report mean Average Precision (mAP) as an evaluation metric, calculating mAP at different temporal intersection over union (tIoU) thresholds.
            For THUMOS14, the tIoU thresholds are set at intervals of 0.1 from 0.3 to 0.7, while for ActivityNet~v1.3 and HACS-Segment, they range from 0.5 to 0.95 at intervals of 0.05.

        \noindent \textbf{Implementation Details.}
            \label{Sec:details}
            We employ I3D~\cite{I3D}, VideoMAE\hspace{-0.1em}~\cite{videomae} and CoCA~\cite{coca} as our video encoders to extract visual features following the existing methods \cite{2EffPrompt,3STALE,9TiFAD,8T3AL}.
            For THUMOS14, video features are extracted utilizing a sliding window of 16 frames with a stride of 4, while for ActivityNet~v1.3 and HACS-Segment, we extract 16 frames as a segment with a stride of 16.
            For the text encoding, we employ the frozen pre-trained text encoders, such as CLIP~\cite{CLIP} and CoCA~\cite{coca}.
            We follow the prompt design of Ti-FAD~\cite{9TiFAD}, where the input prompt consists solely of the action class name.
            We train our model for 25 epochs on THUMOS14, and for 15 epochs on ActivityNet~v1.3 and HACS-Segment using Adam with 5 epochs of linear warmup.
            The initial learning rate is 0.0001, 
            updated with 
            multi-step decay for THUMOS14 and 
            cosine annealing \cite{cosine_schedule} for ActivityNet~v1.3 and HACS-Segment, 
            respectively.
            All experiments are performed on a single NVIDIA A100 GPU.

    \subsection{Main Results}

        \noindent \textbf{In-distribution Evaluation.}
        Table~\ref{tab:thumos} shows a comparison of state-of-the-art ZSTAD methods on THUMOS14 and ActivityNet v1.3 under the in-distribution setting, where training and test classes are drawn from the same dataset but are disjoint.
        In the no external information scenario, our TF-CADE outperforms the existing models across both split settings (50\%-50\% and 75\%-25\%) with both I3D~\cite{I3D} and CoCA~\cite{coca} features. 
        Notably, under the 50\%-50\% split on THUMOS14, our method significantly outperforms the previous best by large margins.
        On ActivityNet~v1.3, our model also improves the average mAP compared to previous methods.
        These results demonstrate the effectiveness of our method, even without relying on any external classifier.
        

    \begin{table}[!t]

\centering
\caption{\textbf{Performance comparison with the state-of-the-art methods on THUMOS14 under zero-shot cross-dataset generalization setting.} All methods are trained only on ActivityNet~v1.3 and directly evaluated on THUMOS14 without any fine-tuning.}
\vspace{-3pt}
\renewcommand{\arraystretch}{1.1}
\resizebox{\linewidth}{!}{
\begin{tabular}{clcccccccc}
\toprule[1.5pt]
\multirow{2.5}{*}{\shortstack{Evaluation\\Setting}} 
& \multicolumn{1}{l}{\multirow{2.5}{*}{Model}} 
& \multicolumn{2}{c}{Feature} 
& \multicolumn{6}{c}{THUMOS14} \\
\noalign{\smallskip}
\cline{3-10}
\noalign{\smallskip}
 & & Video & Text & 0.3 & 0.4 & 0.5 & 0.6 & 0.7 & Avg. \\
\noalign{\smallskip}
\hline
\hline
\noalign{\smallskip}
\multirow{5}{*}{\shortstack{50\%-50\%}}  
& STALE \cite{3STALE}  & I3D \cite{I3D} & CLIP-B \cite{CLIP} & 1.3 & 0.7 & 0.6 & 0.6 & 0.4 & 0.7 \\
& EffPrompt \cite{2EffPrompt}  & I3D \cite{I3D} & CLIP-B \cite{CLIP} & 5.4 & 4.4 & 3.5 & 2.7 & 1.9 & 3.6  \\
& T3AL \cite{8T3AL}  & CoCA \cite{coca} & CoCA \cite{coca} & 20.7 & 14.3 & 8.9 & 5.3 & 2.7 & 10.4  \\
& Ti-FAD \cite{9TiFAD}  & I3D \cite{I3D} & CLIP-B \cite{CLIP} & 19.3 & 15.8 & 11.7 & 7.7 & 4.1 & 11.7  \\
& \cellcolor[gray]{0.92}\textbf{TF-CADE (Ours)} & \cellcolor[gray]{0.92}I3D \cite{I3D} & \cellcolor[gray]{0.92}CLIP-B \cite{CLIP} & \cellcolor[gray]{0.92}\textbf{40.5} & \cellcolor[gray]{0.92}\textbf{34.1} & \cellcolor[gray]{0.92}\textbf{26.5} & \cellcolor[gray]{0.92}\textbf{17.2} & \cellcolor[gray]{0.92}\textbf{8.9} & \cellcolor[gray]{0.92}\textbf{28.2} \\
\noalign{\smallskip}
\hline
\noalign{\smallskip}
\multirow{5}{*}{\shortstack{75\%-25\%}}
& STALE \cite{3STALE}  & I3D \cite{I3D} & CLIP-B \cite{CLIP} & 0.5 & 0.3 & 0.2 & 0.2 & 0.2 & 0.3  \\
& EffPrompt \cite{2EffPrompt}  & I3D \cite{I3D} & CLIP-B \cite{CLIP} & 7.1 & 5.9 & 4.5 & 3.4 & 2.2 & 4.6 \\
& T3AL \cite{8T3AL}  & CoCA \cite{coca}  & CoCA \cite{coca}  & 19.2 & 12.7 & 7.4 & 4.4 & 2.2 & 9.2  \\
& Ti-FAD \cite{9TiFAD}  & I3D \cite{I3D} & CLIP-B \cite{CLIP}  & 21.9 & 17.6 & 12.7 & 8.5 & 4.4 & 13.0 \\
& \cellcolor[gray]{0.92}\textbf{TF-CADE (Ours)} & \cellcolor[gray]{0.92}I3D \cite{I3D} & \cellcolor[gray]{0.92}CLIP-B \cite{CLIP} & \cellcolor[gray]{0.92}\textbf{41.8} & \cellcolor[gray]{0.92}\textbf{34.9} & \cellcolor[gray]{0.92}\textbf{27.1} & \cellcolor[gray]{0.92}\textbf{17.6} & \cellcolor[gray]{0.92}\textbf{9.2} & \cellcolor[gray]{0.92}\textbf{26.1} \\
\noalign{\smallskip}
\hline
\noalign{\smallskip}
\multirow{3}{*}{\shortstack{0\%-100\%}}
& T3AL \cite{8T3AL}  & CoCA \cite{coca}  & CoCA \cite{coca} & 18.3 & 12.9 &  8.7 & 5.3 & 2.9 & 9.6 \\
& Ti-FAD \cite{9TiFAD}  & I3D \cite{I3D} & CLIP-B \cite{CLIP}  & 17.9 & 15.0 & 11.1 & 7.4 & 4.0 & 11.1 \\
& \cellcolor[gray]{0.92}\textbf{TF-CADE (Ours)} & \cellcolor[gray]{0.92}I3D \cite{I3D} & \cellcolor[gray]{0.92}CLIP-B \cite{CLIP} & \cellcolor[gray]{0.92}\textbf{42.8} & \cellcolor[gray]{0.92}\textbf{36.5} & \cellcolor[gray]{0.92}\textbf{28.6} & \cellcolor[gray]{0.92}\textbf{19.0} & \cellcolor[gray]{0.92}\textbf{10.1} & \cellcolor[gray]{0.92}\textbf{27.4}\\

\bottomrule[1.5pt]

\end{tabular}}
\vspace{-3pt}
\label{tab:cross_thumos}
\vspace{-18pt}
\end{table}

\begin{table*}[!t]
\caption{\textbf{Performance comparison with the state-of-the-art methods under zero-shot cross-dataset generalization setting.} All models are trained on only the training dataset and directly evaluated on the target datasets without any fine-tuning.}
\vspace{-3pt}
\centering
\renewcommand{\arraystretch}{1.1}

\begin{subtable}{0.82\linewidth}
\centering

\resizebox{\linewidth}{!}{
\begin{tabular}{clcccccccccccccc}
\toprule[1.5pt]
\multirow{2.5}{*}{Training Dataset} 
& \multicolumn{1}{l}{\multirow{2.5}{*}{Model}} 
& \multicolumn{2}{c}{Feature} 
& & \multicolumn{6}{c}{THUMOS14} 
& & \multicolumn{4}{c}{ActivityNet~v1.3} \\
\noalign{\smallskip}
\cline{3-4}
\cline{6-11}
\cline{13-16}
\noalign{\smallskip}
 & & Video & Text & & 0.3 & 0.4 & 0.5 & 0.6 & 0.7 & Avg. & & 0.5 & 0.75 & 0.95 & Avg.\\
\noalign{\smallskip}
\hline
\hline
\noalign{\smallskip}
\multirow{2}{*}{\shortstack{HACS-Segment}}  
& Ti-FAD \cite{9TiFAD}  & VideoMAE \cite{videomae} & CLIP-B \cite{CLIP} & & 32.5 & 30.1 & 26.5 & 21.6 & 14.3 & 25.0 & & 43.1 & \textbf{22.7} & 2.3 & 24.1 \\
& \cellcolor[gray]{0.92}\textbf{TF-CADE (Ours)} & \cellcolor[gray]{0.92}VideoMAE \cite{videomae} & \cellcolor[gray]{0.92}CLIP-B \cite{CLIP} & \cellcolor[gray]{0.92} & \cellcolor[gray]{0.92}\textbf{41.8} & \cellcolor[gray]{0.92}\textbf{37.9} & \cellcolor[gray]{0.92}\textbf{32.4} & \cellcolor[gray]{0.92}\textbf{25.2} & \cellcolor[gray]{0.92}\textbf{16.2} & \cellcolor[gray]{0.92}\textbf{30.7} & \cellcolor[gray]{0.92} & \cellcolor[gray]{0.92}\textbf{46.2} & \cellcolor[gray]{0.92}22.2 & \cellcolor[gray]{0.92}\textbf{3.1} & \cellcolor[gray]{0.92}\textbf{24.8}\\

\bottomrule[1.5pt]
\end{tabular}}
\caption{Trained on HACS-Segment~\cite{hacs}}
\label{tab:cross_hacs}
\vspace{-3pt}
\end{subtable}

\vspace{5pt}

\begin{subtable}{0.82\linewidth}
\centering

\resizebox{\linewidth}{!}{
\begin{tabular}{clcccccccccccccc}
\toprule[1.5pt]
\multirow{2.5}{*}{Training Dataset} 
& \multicolumn{1}{l}{\multirow{2.5}{*}{Model}} 
& \multicolumn{2}{c}{Feature} & 
& \multicolumn{6}{c}{THUMOS14} &
& \multicolumn{4}{c}{HACS-Segment} \\
\noalign{\smallskip}
\cline{3-4}
\cline{6-11}
\cline{13-16}
\noalign{\smallskip}
 & & Video & Text & & 0.3 & 0.4 & 0.5 & 0.6 & 0.7 & Avg. & & 0.5 & 0.75 & 0.95 & Avg.\\
\noalign{\smallskip}
\hline
\hline
\noalign{\smallskip}
\multirow{2}{*}{\shortstack{ActivityNet~v1.3}}
& Ti-FAD \cite{9TiFAD}  & VideoMAE \cite{videomae} & CLIP-B \cite{CLIP} & & 21.8 & 17.9 & 13.3 & 8.5 & 4.3 & 13.2 & & 22.0 & 7.0 & 0.6 & 9.7 \\
& \cellcolor[gray]{0.92}\textbf{TF-CADE (Ours)} & \cellcolor[gray]{0.92}VideoMAE \cite{videomae} & \cellcolor[gray]{0.92}CLIP-B \cite{CLIP} & \cellcolor[gray]{0.92} & \cellcolor[gray]{0.92}\textbf{36.9} & \cellcolor[gray]{0.92}\textbf{30.1} & \cellcolor[gray]{0.92}\textbf{22.2} & \cellcolor[gray]{0.92}\textbf{13.6} & \cellcolor[gray]{0.92}\textbf{6.7} & \cellcolor[gray]{0.92}\textbf{21.9} & \cellcolor[gray]{0.92} & \cellcolor[gray]{0.92}\textbf{30.0} & \cellcolor[gray]{0.92}\textbf{10.2} & \cellcolor[gray]{0.92}\textbf{1.0} & \cellcolor[gray]{0.92}\textbf{13.6}\\
\bottomrule[1.5pt]

\end{tabular}}
\caption{Trained on ActivityNet~v1.3~\cite{ActivityNet}}
\label{tab:cross_anet}
\end{subtable}

\vspace{-5pt}
\label{tab:cross}
\vspace{-11pt}
\end{table*}




        \noindent \textbf{Cross-dataset Generalization.}
            We evaluate the generalization ability of TF-CADE in the cross-dataset setting, where models are trained on ActivityNet~v1.3 and tested on THUMOS14.
            Table~\ref{tab:cross_thumos} shows the performance on THUMOS14 under three different splits.
            The 50\%-50\% and 75\%-25\% settings are included solely to enable fair comparison with existing evaluation protocols.
            TF-CADE consistently outperforms prior works across all splits. 
            Notably, under the most challenging 0\%-100\% setting, TF-CADE outperforms both T3AL~\cite{8T3AL} and Ti-FAD~\cite{9TiFAD} by large margins.
            In Table~\ref{tab:cross}, we further evaluate the generalization ability, including the larger-scale dataset HACS-Segment by comparing it with THUMOS14 and ActivityNet~v1.3.
            For fair cross-dataset evaluation, we use VideoMAE~\cite{videomae} features that are available for all three datasets under same feature extraction setting.
            As demonstrated in Table~\ref{tab:cross}, TF-CADE achieves consistent improvements over Ti-FAD~\cite{9TiFAD} on all datasets, indicating strong robustness in diverse cross-dataset scenarios.

    \subsection{Further Analysis}
    \noindent \textbf{Ablation Study on Cross-modal Update.}
    As discussed in Sec.~\ref{sec:baseline}, we adopt the cross-modal adaptation proposed in Ti-FAD~\cite{9TiFAD} as our baseline.
    This baseline updates both text and video features through a bi-directional cross-attention mechanism, where each modality alternately serves as the query.
    As shown in Table~\ref{tab:baseline}, the performance difference among using text, video, or both as the query is marginal.
    Therefore, in our framework, we adopt a simplified design that updates only the video features through cross-attention.

    \noindent \textbf{Component Analysis.}
    Table~\ref{tab:component} shows an ablation study of each component in TF-CADE on THUMOS14.
    The second row denotes that the model is trained using the foreground-based video-level similarity score $S_{fg}$ produced by our ACA, which yields marginal improvement when used alone.
    When applying CCR alone, the model exhibits larger performance gain by suppressing irrelevant classes.
    Finally, combining both $\mathcal{L}_{video}$ and CCR shows best performance, demonstrating that two components are complementary, as $S_{fg}$ trained with $\mathcal{L}_{video}$ provides a global prior that improves snippet-level classification by re-weighting confidence scores.



    \noindent \textbf{Effectiveness of Smoothing Filter.}
    To validate the effectiveness of applying a Gaussian smoothing filter to the temporal action certainty map, we compare performance under the zero-shot cross-dataset setting.
    Table~\ref{tab:filter} shows that applying the Gaussian smoothing filter leads to a clear improvement, demonstrating its effectiveness in reducing the impact of noise and overly sharp peaks in action segments.
    This smoothing process enables the model to concentrate on the foreground regions, leading to better generalization.

\begin{table}[!t]
    \caption{\textbf{Comparison of cross-attention configurations.} The table shows how different modality update choices affect performance and support our simplified video-only update design.}
    \vspace{-3pt}
    \centering
    \small
    \resizebox{0.77\linewidth}{!}{
    \begin{tabular}{c c c cccc}
    \toprule[1.5pt]
     \multirow{2.5}{*}{Method} & \multirow{2.5}{*}{Text} & \multirow{2.5}{*}{Video} & \multicolumn{4}{c}{mAP@tIoU (\%)} \\ 
     \noalign{\smallskip}
     \cline{4-7}
     \noalign{\smallskip}
     & & & 0.3 & 0.5 & 0.7 & Avg. \\
     \noalign{\smallskip}
     \hline
     \hline
     \noalign{\smallskip}
    \multirow{3}{*}{Baseline} & \checkmark &  & 20.1 & 15.8 & 8.0 & 14.9 \\ 
     &  & \checkmark & 21.4 & 17.0 & 8.7 & 16.0 \\ 
     & \checkmark & \checkmark & 22.3 & 17.4 & 8.4 & 16.3 \\
    \noalign{\smallskip}
    \hline
    \noalign{\smallskip}
    \multirow{3}{*}{Ours} & \checkmark & & 27.3 & 21.2 & 9.8 & 19.9 \\ 
     & \cellcolor[gray]{0.92} & \cellcolor[gray]{0.92}\checkmark & \cellcolor[gray]{0.92}\textbf{28.6} & \cellcolor[gray]{0.92}\textbf{22.6} & \cellcolor[gray]{0.92}\textbf{10.7} & \cellcolor[gray]{0.92}\textbf{21.1} \\
     & \checkmark & \checkmark & 27.1 & 20.6 & 9.3 & 19.3 \\ 
    \bottomrule[1.5pt]
    \end{tabular}
    }
    \vspace{-2pt}
    \label{tab:baseline}
    \vspace{-5pt}
\end{table}



\begin{table}[!t]
    \centering
    \caption{\textbf{Component analysis on THUMOS14.} The average mAP is calculated to evaluate performance at all tIoU thresholds.}
    \vspace{-3pt}
    \resizebox{0.79\linewidth}{!}{
    \begin{tabular}{lcccccc}
    \toprule[1.5pt]
    \multirow{2.5}{*}{Method} & \multicolumn{6}{c}{mAP@tIoU(\%)} \\
    \noalign{\smallskip}
    \cline{2-7}
    \noalign{\smallskip}
    & 0.3 & 0.4 & 0.5 & 0.6 & 0.7 & Avg. \\
    \noalign{\smallskip}
    \hline
    \hline
    \noalign{\smallskip}
    \text{Baseline} & 21.4 & 19.6 & 17.0 & 13.4 & 8.7 & 16.0 \\
    $+ \mathcal{L}_{video}$ & 22.0 & 20.2 & 17.4 & 13.6 & 8.6 & 16.4 \\
    $+ \text{CCR}$ & 27.1 & 24.9 & 21.0 & 16.0 & 9.8 & 19.7\\
    \rowcolor[gray]{0.92} $+ \mathcal{L}_{video}$ \& CCR & \textbf{28.6} & \textbf{26.2} & \textbf{22.6} & \textbf{17.1} & \textbf{10.7} & \textbf{21.1} \\
    \bottomrule[1.5pt]
    \end{tabular}}
    \vspace{-2pt}
    \label{tab:component}
    \vspace{-5pt}
\end{table}

    \begin{table}[!t]
\caption{\textbf{Effectiveness of smoothing filter under zero-shot cross-dataset generalization setting.} The models are trained on ActivityNet~v1.3 and evaluated on THUMOS14.}
\vspace{-3pt}
\centering
\resizebox{0.67\linewidth}{!}{
\begin{tabular}{lcccccc}
\toprule[1.5pt]
\noalign{\smallskip}
\multirow{2.5}{*}{Method} & \multicolumn{6}{c}{mAP@tIoU(\%)} \\
\noalign{\smallskip}
\cline{2-7}
\noalign{\smallskip}
& 0.3 & 0.4 & 0.5 & 0.6 & 0.7 & Avg. \\
\noalign{\smallskip}
\hline
\hline
\noalign{\smallskip}
 \text{w/o filter} & 35.7 & 28.6 & 21.7 & 13.5 & 6.7 & 21.3 \\
\rowcolor[gray]{0.92}\text{w/ filter} & \textbf{42.8} & \textbf{36.5} & \textbf{28.6} & \textbf{19.0} & \textbf{10.1} & \textbf{27.4} \\
\bottomrule[1.5pt]
\end{tabular}}
\vspace{-2pt}
\label{tab:filter}
\vspace{-11pt}
\end{table}

    \begin{table*}[!t]
    \caption{\textbf{Ablation studies of ACA on THUMOS14 in the 50\%-50\% setting.}}
    \vspace{-3pt}
    \centering
        \begin{subtable}{0.25\linewidth}
            \centering

            \resizebox{\linewidth}{!}{
                \begin{tabular}{ccccc}
                \toprule[1.5pt]
                \multirow{2.5}{*}{Type} & \multicolumn{4}{c}{mAP@tIoU (\%)} \\
                \noalign{\smallskip}
                \cline{2-5}
                \noalign{\smallskip}
                & 0.3 & 0.5 & 0.7 & Avg. \\
                \noalign{\smallskip}
                \hline \hline
                \noalign{\smallskip}
                Mean pooling & 26.4 & 20.2 & 9.0 & 18.9 \\
                 \rowcolor[gray]{0.92}\textbf{Certainty} & \textbf{28.6} & \textbf{22.6} & \textbf{10.7} & \textbf{21.1} \\
                \bottomrule[1.5pt]
                \end{tabular}
            }
            \caption{Aggregation method}
            \label{tab:design}
        \end{subtable}
        \hfill
        \begin{subtable}{0.21\linewidth}
            \centering

            \resizebox{\linewidth}{!}{
                \begin{tabular}{ccccc}
                \toprule[1.5pt]
                \multirow{2.5}{*}{Type} & \multicolumn{4}{c}{mAP@tIoU (\%)} \\
                \noalign{\smallskip}
                \cline{2-5}
                \noalign{\smallskip}
                & 0.3 & 0.5 & 0.7 & Avg. \\
                \noalign{\smallskip}
                \hline \hline
                \noalign{\smallskip}
                $m_{\text{max}}^{(l)}$ & 28.0 & 21.9 & 10.9 & 20.7  \\
                $m_{\text{filter}}^{(l)}$ & 26.6 & 20.9 & 10.2 & 19.6\\
                 \rowcolor[gray]{0.92}\textbf{$m_{\text{max}}^{(l)}$ + $m_{\text{filter}}^{(l)}$} & \textbf{28.6} & \textbf{22.6} & \textbf{10.7} & \textbf{21.1}  \\
                 
                \bottomrule[1.5pt]
                \end{tabular}
            }
            \caption{Action certainty map}
            \label{tab:hardneg}
        \end{subtable}
        \hfill
        \hfill
        \begin{subtable}{0.26\linewidth}
            \centering

            \resizebox{\linewidth}{!}{
                \begin{tabular}{ccccc}
                \toprule[1.5pt]
                \multirow{2.5}{*}{Type} & \multicolumn{4}{c}{mAP@tIoU (\%)} \\
                \noalign{\smallskip}
                \cline{2-5}
                \noalign{\smallskip}
                & 0.3 & 0.5 & 0.7 & Avg. \\
                \noalign{\smallskip}
                \hline \hline
                \noalign{\smallskip}
                 Learnable (1D conv) & 28.7 & 21.7 & 9.9 & 20.5 \\
                Gaussian + Learnable & 27.5 & 21.3 & 10.5 & 20.2\\
                 \rowcolor[gray]{0.92}\textbf{Fix (Gaussian)} & \textbf{28.6} & \textbf{22.6} & \textbf{10.7} & \textbf{21.1} \\
                \bottomrule[1.5pt]
                \end{tabular}
            }
            \caption{Smoothing strategy}
            \label{tab:pos_sample}
        \end{subtable}
        \hfill
        \begin{subtable}{0.19\linewidth}
            \centering

            \resizebox{\linewidth}{!}{
                \begin{tabular}{ccccc}
                \toprule[1.5pt]
                \multirow{2.5}{*}{Type} & \multicolumn{4}{c}{mAP@tIoU (\%)} \\
                \noalign{\smallskip}
                \cline{2-5}
                \noalign{\smallskip}
                & 0.3 & 0.5 & 0.7 & Avg. \\
                \noalign{\smallskip}
                \hline \hline
                \noalign{\smallskip}
                 \rowcolor[gray]{0.92}\textbf{$\sigma=1$} & \textbf{28.6} & \textbf{22.6} & \textbf{10.7} & \textbf{21.1}\\
                $\sigma=2$ & 28.8 & 22.1 & 10.1 & 20.7\\
                \textbf{$\sigma=3$} & 28.7 & 22.3 & 10.2 & 20.8 \\
                \bottomrule[1.5pt]
                \end{tabular}
            }
            \caption{$\sigma$ in Gaussian kernel}
            \label{tab:neg_sample}
        \end{subtable}
        \hfill
        \\

    \vspace{-3pt}
    \vspace{-3pt}
    \label{tab:ablation_acc}
    \vspace{-11pt}
\end{table*}

            

\begin{figure}[!t]
        \centering
        \begin{subfigure}[t]{0.8\linewidth}
            \centering
            \includegraphics[width=\linewidth]{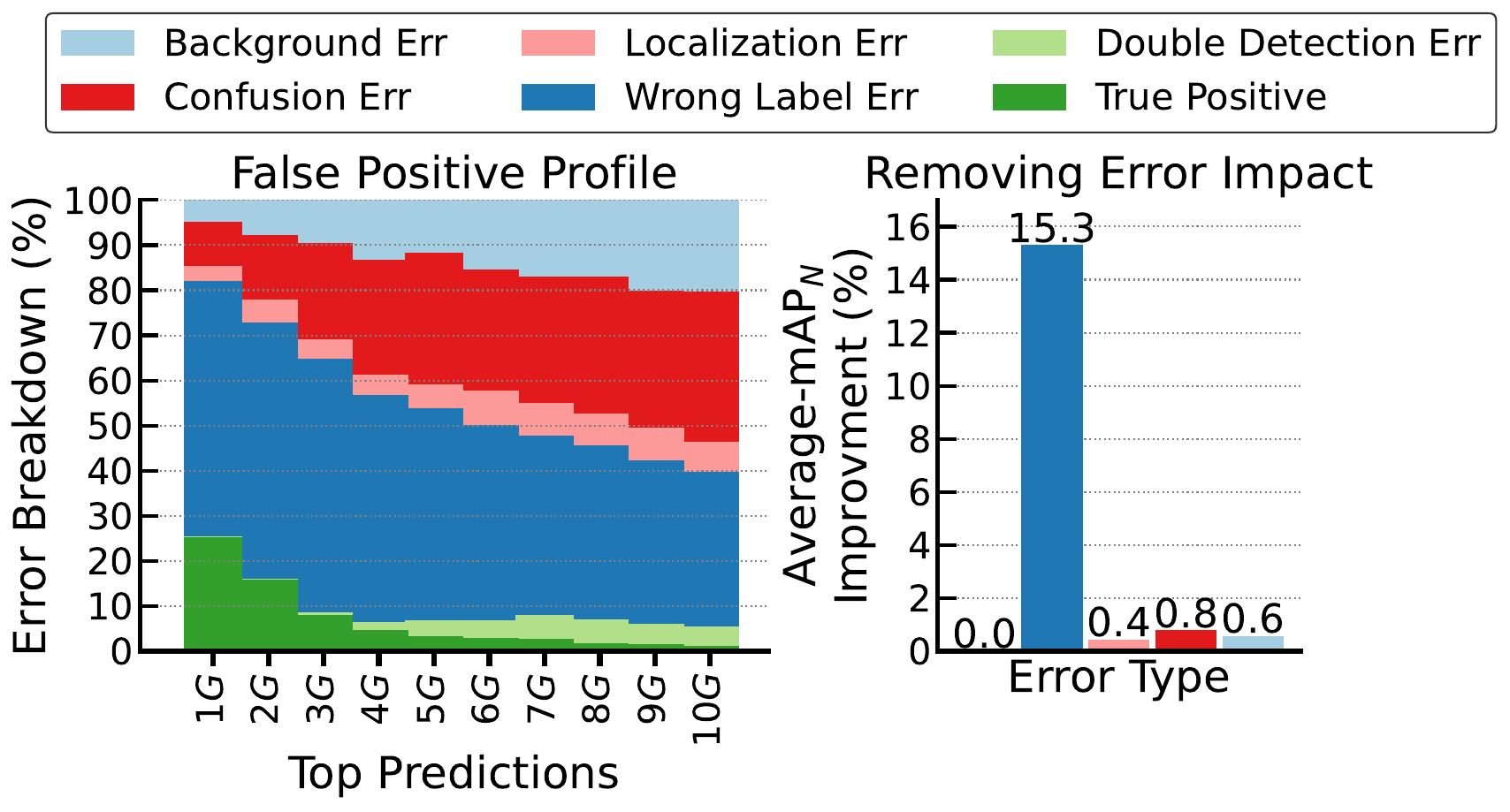}
            \vspace{-15pt}
            \caption{Ti-FAD~\cite{9TiFAD}}
        \end{subfigure}
        \begin{subfigure}[t]{0.8\linewidth}
            \centering
            \includegraphics[width=\linewidth]{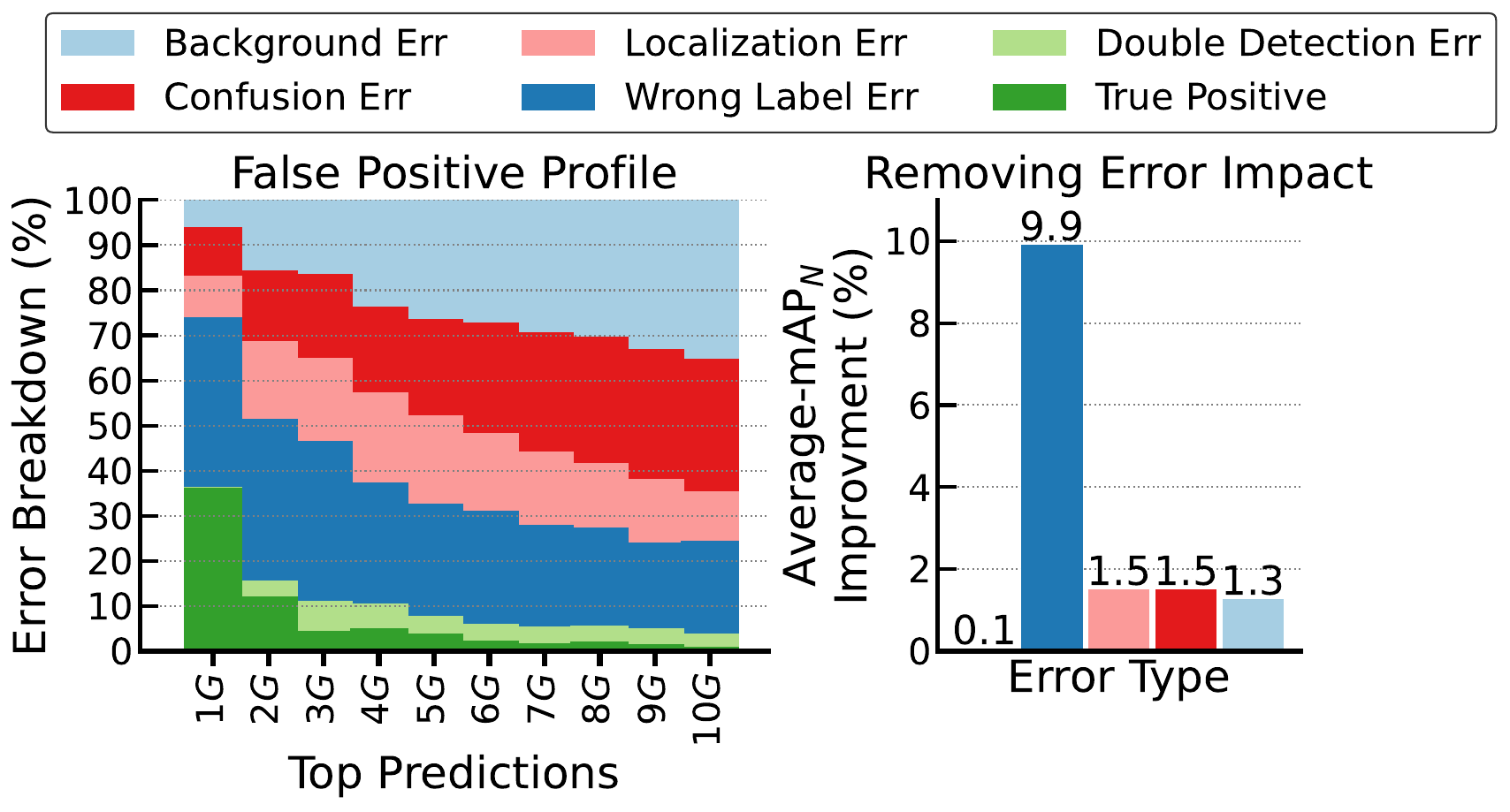} 
            \vspace{-15pt}
            \caption{TF-CADE (Ours)}
        \end{subfigure}
        \vspace{-5pt}
        \caption{\textbf{False positive analysis on THUMOS14 using DETAD~\cite{DETAD}.} Compared to Ti-FAD, our TF-CADE reduces the dependency on the Wrong Label Error.}

        \label{fig:false_positive}
        \vspace{-17pt}
    \end{figure}

    \noindent \textbf{False Positive Analysis.}    Fig.~\ref{fig:false_positive} illustrates a false positive analysis on THUMOS14 using DETAD~\cite{DETAD}.
    The analysis contains error types such as wrong label errors.
    Compared to Ti-FAD, our TF-CADE exhibits a notable reduction in wrong label errors.  
    This reduction indicates that TF-CADE more effectively injects text features into the detection process, leading to improved action classification.

\begin{figure}[!t]
        \centering
        \vspace{-4pt}
        \includegraphics[width=0.85\linewidth]{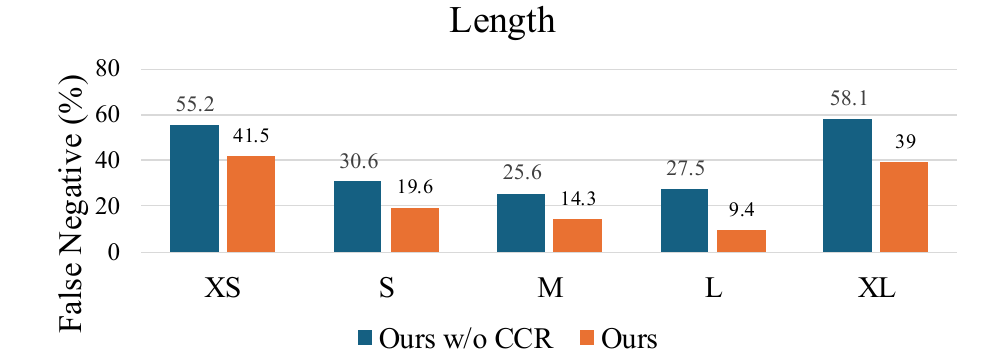}
        \vspace{-4pt}
        \caption{\textbf{False negative analysis on THUMOS14 using DETAD~\cite{DETAD}.} CCR reduces false negatives consistently across actions of different temporal durations, especially for extremely short (XS) and long (XL) actions.}
        \label{fig:false_negative}
        \vspace{-5pt}
    \end{figure}

    \noindent \textbf{False Negative Analysis.}
    Fig.~\ref{fig:false_negative} shows a false negative analysis on THUMOS14 using DETAD~\cite{DETAD} across different action lengths.
    Without CCR, the model suffers from high false negatives, particularly for extremely short (XS) and long (XL) actions, where snippet-level classification alone fails to capture the complete temporal context.
    By introducing CCR, which re-weights snippet-level confidence using the video-level certainty score obtained from the action concentrate aggregation, TF-CADE effectively suppresses false negatives in these challenging cases.
    This demonstrates that video-level score re-weighting provides a complementary cue to refine snippet-wise predictions, enabling more reliable detection across varying temporal durations.
    

    \noindent \textbf{Ablation and Design Analysis of ACA.}
    Table~\ref{tab:ablation_acc} shows how different architectural choices in ACA, certainty maps, smoothing strategies, and kernel parameters.
    To validate the effectiveness of ACA, we compare our ACA with mean pooling, which is the simplest and most commonly used temporal aggregation method.
    Table~\ref{tab:ablation_acc}~(a) shows that ACA consistently outperforms the baseline across all tIoU thresholds.
    This result highlights the importance of aggregating foreground regions into video-level representations.
    Table~\ref{tab:ablation_acc}~(b) illustrates the types of action certainty maps.
    Combining $m_{\text{max}}^{(l)}$ and $m_{\text{filter}}^{(l)}$ yields the best performance, compared to using either map alone, indicating that their integration provides a more reliable aggregation of action certainty.
    Table~\ref{tab:ablation_acc}~(c) compares three smoothing strategies applied to the action certainty map $m_{\text{filter}}^{(l)}$.
    The first row uses a learnable 1D convolution filter, while the second initializes a learnable filter with a Gaussian kernel.
    The third row applies a fixed Gaussian smoothing filter without any learnable parameters.
    Among the three, the fixed Gaussian filter yields the highest performance, and thus we adopt this smoothing strategy in our final model.
    Furthermore, we analyze the hyperparameter $\sigma$ in Gaussian kernel in Table~\ref{tab:ablation_acc}~(d).
    Performance remains stable across different sigma values, demonstrating that our method is robust to the choice of $\sigma$.

\begin{figure}[!t]
    \centering
    \vspace{-3pt}
    \includegraphics[width=0.8\linewidth]{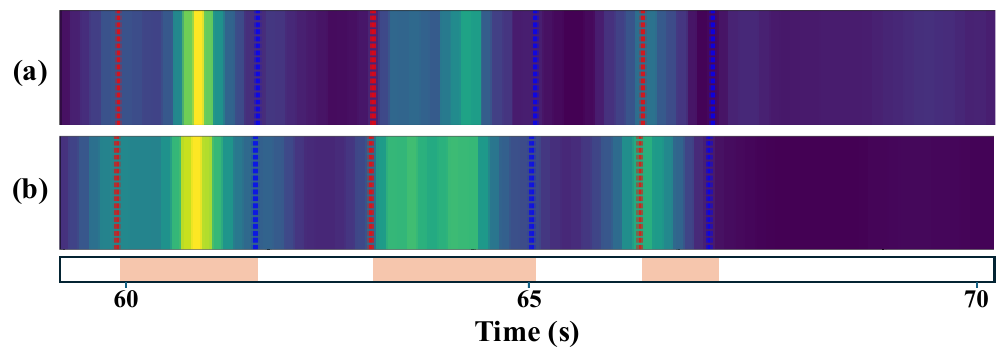}
    \vspace{-7pt}
    \caption{\textbf{Visualization of Temporal Action Certainty Maps on THUMOS14.}
    (a) Certainty map $m_{\text{max}}^{(l)}$ without smoothing. 
    (b) Filtered Certainty map $m_{\text{filter}}^{(l)}$ by Gaussian kernel. 
    The bar below plot indicates the ground-truth action segments.
    }
    \label{fig:certainty_map}
    \vspace{-15pt}
    \end{figure}

    \noindent \textbf{Visualization of Certainty Map in ACA.}
    Fig.~\ref{fig:certainty_map} visualizes the temporal action certainty maps produced by our ACA.
    In Fig.~\ref{fig:certainty_map}~(a), the certainty map based on $m_{\text{max}}^{(l)}$ exhibits sharp and fragmented activations.
    After applying the Gaussian filter, as shown in Fig.~\ref{fig:certainty_map}~(b), the certainty distribution of $m_{\text{filter}}^{(l)}$ becomes smoother and better highlights foreground action regions.
    This smoothing process reduces spurious peaks and preserves scores within action segments.
    ACA allows the model to focus on action-relevant regions while down-weighting ambiguous or background snippets.
    The refined certainty map provides a stable and reliable confidence basis for video-level re-weighting in CCR.
    Additional visualizations and analyses are presented in the Supp.

\section{Conclusion}
\label{sec:conclusion}
In this paper, we address the issue of text-irrelevant predictions in ZSTAD. 
We propose TF-CADE, a novel framework that explicitly aligns textual information with action-relevant foreground regions. 
Our proposed Action~Concentrate Aggregation~(ACA) enhances text-video alignment by aggregating temporally informative segments into a foreground-weighted video embedding, while our Certainty-based~Confidence~Re-weighting~(CCR) refines snippet-level predictions by leveraging video-level similarity score during inference. 
Extensive experiments on THUMOS14, ActivityNet~v1.3, and HACS-Segment demonstrate that TF-CADE achieves state-of-the-art performance under both in-distribution and cross-dataset generalization settings, validating its effectiveness and robustness.
We believe that our findings provide new insights into leveraging foreground concentration for more semantically grounded text-video alignment in ZSTAD.\hspace{-0.1em}
\newpage
\section*{Acknowledgement}
This work was supported by the Institute of Information \& Communications Technology Planning \& Evaluation (IITP) grant, funded by 
the Korea government (MSIT) (No. RS-2019-II190079, Artificial Intelligence Graduate School Program (Korea University), No. RS-2024-00457882, AI Research Hub Project, No. RS-2025-02304828, Artificial Intelligence Star Fellowship Support Program to Nurture the Best Talents, and No. RS-2022-II220984, Development of Artificial Intelligence Technology for Personalized Plug-and-Play Explanation and Verification of Explanation).
{
    \small
    \bibliographystyle{ieeenat_fullname}
    \bibliography{main}
}

\end{document}